\documentclass[final,3p,times]{elsarticle}

\usepackage{multirow}
\usepackage{booktabs}
\usepackage{tabularx}
\usepackage{xcolor}         
\usepackage{algorithm}
\usepackage{algorithmic}
\usepackage{epsfig}
\usepackage{graphicx}
\usepackage{amsmath}
\usepackage{amssymb}
\usepackage{bm}
\usepackage{arydshln}
\usepackage{wrapfig}
\usepackage{float}          
\usepackage{placeins}       
\usepackage{pifont}
\usepackage{newunicodechar}
\newunicodechar{✓}{\ding{51}}
\definecolor{iccvblue}{rgb}{0.21,0.49,0.74}
\definecolor{lightyellow}{rgb}{0.9, 0.85, 0.55}

\journal{Pattern Recognition}

\begin{document}

\begin{frontmatter}



\title{Noise-Robust Tiny Object Localization with Flows}


\author{
Huixin Sun\textsuperscript{1},
Linlin Yang\textsuperscript{2},
Ronyu Chen\textsuperscript{3}, \\
Kerui Gu\textsuperscript{3}, 
Baochang Zhang\textsuperscript{1},
Angela Yao\textsuperscript{3},
Xianbin Cao\textsuperscript{1} \\
\textsuperscript{1}Beihang University, China \\
\textsuperscript{2}Communication University of China, China \\ 
\textsuperscript{3}National University of Singapore, Singapore
}

\begin{abstract}
Despite significant advances in generic object detection, a persistent performance gap remains for tiny objects compared to normal-scale objects.
We demonstrate that tiny objects are highly sensitive to annotation noise, where optimizing strict localization objectives risks noise overfitting.
To address this, we propose Tiny Object Localization with Flows (TOLF), a noise-robust localization framework leveraging normalizing flows for flexible error modeling and uncertainty-guided optimization. 
Our method captures complex, non-Gaussian prediction distributions through flow-based error modeling, enabling robust learning under noisy supervision. 
An uncertainty-aware gradient modulation mechanism further suppresses learning from high-uncertainty, noise-prone samples, mitigating overfitting while stabilizing training.
Extensive experiments across three datasets validate our approach's effectiveness.
Especially, TOLF boosts the DINO baseline by 1.2\% AP on the AI-TOD dataset.
\end{abstract}




\end{frontmatter}

\section{Introduction}
\label{sec:intro}
Recent progress in deep neural networks (DNNs)~\cite{he2015convolutional} has significantly advanced object detection field~\cite{lin2014microsoft}.
Despite these advancements, Tiny Object Detection (TOD) remains a highly challenging problem~\cite{yang2022querydet}.
Characterized by extremely limited pixel inputs (less than 16×16 pixels~\cite{wang2021tiny}), tiny objects exhibit severe performance degradation in generic detectors compared to general object detection~\cite{lin2014microsoft}.
For instance, DINO~\cite{zhang2022dino}, a state-of-the-art query-based detector, achieves 37.6\% AP on medium objects but only 9.9\% AP on tiny objects in AI-TOD~\cite{wang2021tiny}.
%
%
The prohibitively low performance falls short of the demands of safety-critical real-world applications, such as traffic management~\cite{long2024interpretable}, driving assistance~\cite{liang2025memory}, and anomaly detection~\cite{zhang2023enhanced}.

The inherently limited pixel inputs of tiny objects constitutes a primary challenge in TOD, which hinders the extraction of sufficient discriminative foreground features~\cite{cao2024visible}.
This challenge is intensified in cluttered environments~\cite{cao2024visible}, where pervasive occlusions, complex background noise~\cite{wang2025fuzzy}, and critically low signal-to-noise ratios induce ambiguity in the feature representation space~\cite{zhou2025spatial}.
Consequently, generic detectors can develop a feature bias towards distinguishing the foreground from background regions that resemble it, which results in missed detections and false positives in TOD.
Recent efforts address these problems by enhancing feature resolution via upsampling or specialized architectures~\cite{lin2017feature}, exploiting contextual information to compensate for limited pixel inputs~\cite{du2023adaptive}, and auxiliary self-reconstruction modules~\cite{cao2024visible} to refine object discrimination.

\begin{figure}[t]
    \centering
    \includegraphics[width=1.0\linewidth]{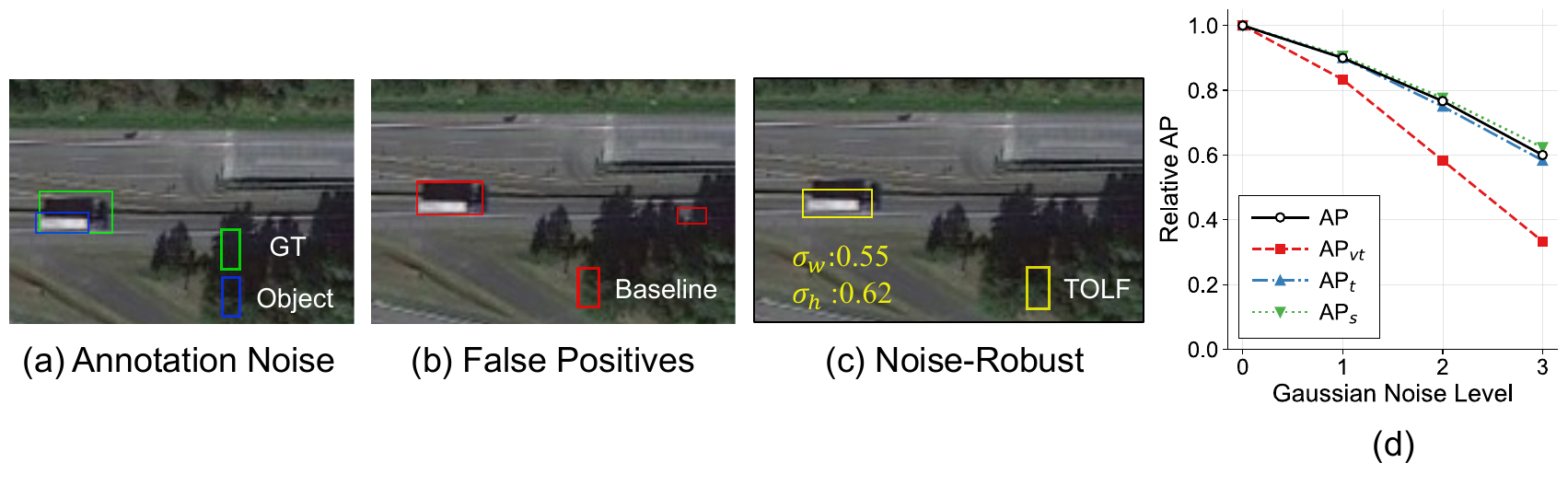}
    \caption{Pathological predictions due to overfitting label noise. (a) Inaccurate ground-truth annotation covering background shadows. (b) Overfitting leads to false positives in background regions that resemble noisy annotations. (c) TOLF exhibits low confidence in uncertain locations and more accurate localization. 
    (d) A noise sensitivity analysis that injects Gaussian noise into training annotations and measures detection performance across object scales. Results reveal tiny objects exhibit the largest degradation.
    Detection performance is evaluated by training a 1$\times$ FCOS detector~\cite{tian2019fcos}.
    The model is trained on the AI-TOD~\cite{wang2021tiny} {\tt trainval} and validated on the AI-TOD {\tt test}. }
    \label{fig:overfit}
\end{figure}

\begin{figure}[t!]
    \centering
    \includegraphics[width=0.8\linewidth]{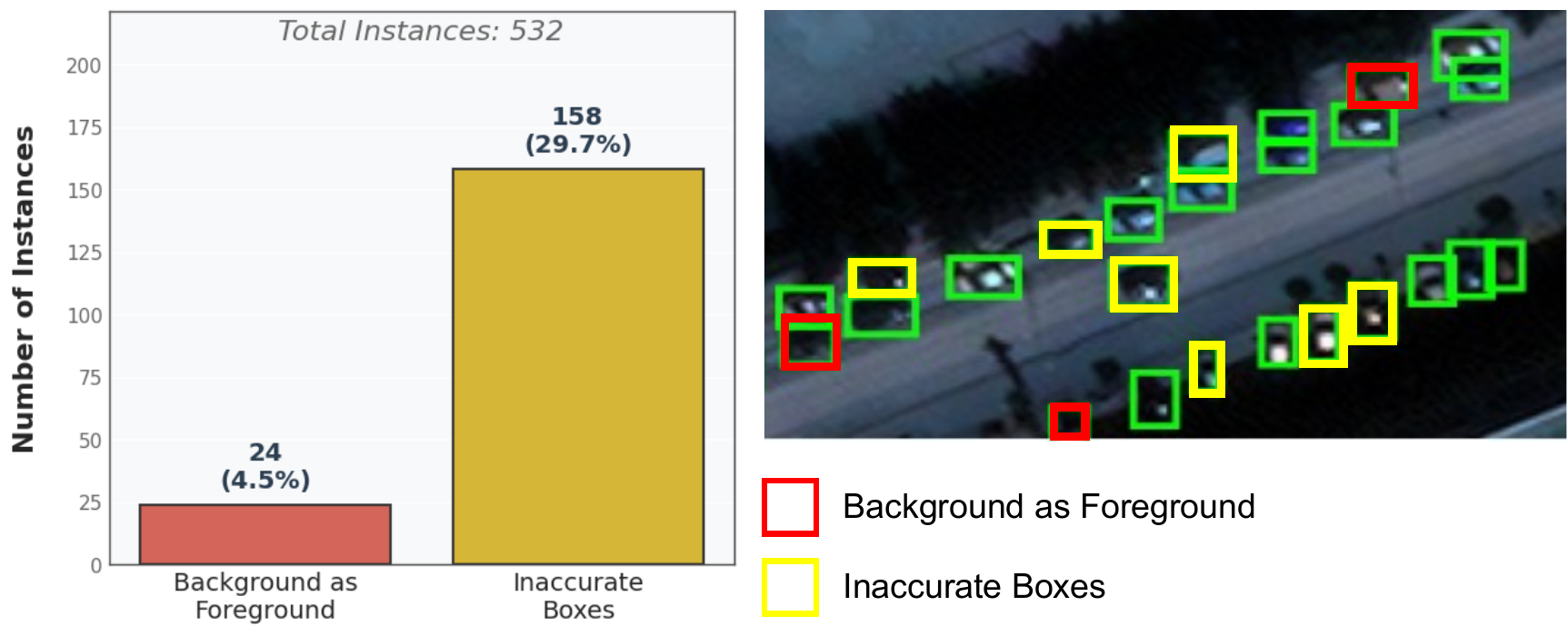}
    \caption{
    Annotation quality statistics based on manual inspection of 532 bounding boxes from 10 AI-TOD {\tt test} images. 
    Red boxes/bars represent background regions mistakenly labeled as foreground, yellow boxes/bars indicate inaccurate or loose bounding boxes.
    The results show that nearly 34.2\% of annotations are noisy.
    }
    \label{fig:noise}
\end{figure}

In this work, we reveal that tiny objects are vulnerable to annotation noise and risks overfitting.
Due to limited resolution and visual ambiguity manual annotations of tiny objects often suffer from labeling inconsistencies, including missed objects, inaccurate bounding boxes, or incorrect classes~\cite{wang2021tiny}.
To quantify the prevalence of annotation noise in real-world tiny object datasets, we manually reviewed 532 bounding boxes across 10 randomly selected images from AI-TOD test in Fig.~\ref{fig:noise}.
Results show that nearly 34.2\% of annotations are noisy.
These errors are exacerbated by the IoU sensitivity of tiny objects, where even a minor deviation can dramatically alter localization quality.
A trivial 2-pixel shift leads to over 20\% IoU drop for a 10×10 object, whereas the same error would only cause ~5\% degradation for a 100×100 object. 
Under such conditions, optimizing for strict localization criteria (e.g., 1.0 IoU) can cause models to overfit annotation noise rather than learning effective regression.
Illustrated in Fig.~\ref{fig:overfit} (b), the overfitting can result in increased false positives in background regions. 
Moreover, we conduct a noise sensitivity analysis to quantify the impact of training-time label noise across object scales.
We inject Gaussian noise with standard deviation \(\sigma \in \{1.0, 2.0, 3.0\}\) pixels into the centers of training bounding boxes and evaluate on clean data. 
As shown in Fig.~\ref{fig:overfit} (d), performance decresed with increasing noise levels across all scales, with the largest degradation for tiny objects.
At \(\sigma=3.0\) pixels, overall AP decreases by 40.0\%, while AP for very tiny and tiny objects decreases by 66.7\%.
This heightened sensitivity to annotation noise highlights the importance of robust localization objectives in tiny object detection.

In light of the analysis, we introduce \textbf{T}iny \textbf{O}bject \textbf{L}ocalization \textbf{F}low (TOLF), a robust localization framework leveraging normalizing flows for flexible prediction distribution modeling, which accounts for uncertainty and label noise.
Unlike conventional uncertainty methods constrained by Gaussian assumptions or fixed priors, TOLF employs invertible normalizing flows to explicitly learn the error distribution between predictions and noisy annotations.
This enables TOLF to capture intricate noise structures, including heavy tails, skewness, and multimodality.
Furthermore, TOLF's loss is uncertainty-aware. 
By adaptively down-weighting noisy examples through uncertainty-based weighting, it suppresses overfitting to annotation errors and maintains stable training under severe noise conditions.
This prevents outliers from dominating the loss landscape.
By unifying flexible error modeling with uncertainty-aware optimization, TOLF provides a principled, data-driven solution that mitigates overfitting at its source,  achieving superior localization robustness and improved accuracy.

To summarize, our main contributions are three-fold:
\begin{enumerate}
\item We demonstrate that tiny object detectors are highly vulnerable to annotation noise, and show that strict localization objectives risk overfitting to noisy labels. To address this, we propose \textbf{TOLF}, a noise-robust localization framework employing flexible distribution modeling.

\item TOLF incorporates a normalizing flow-based error modeling component to capture complex, non-Gaussian error patterns, and  an uncertainty-aware gradient modulation mechanism that adaptively suppresses gradients from high-uncertainty, noise-prone samples.

\item TOLF significantly improves training stability and advances state-of-the-art accuracy for tiny object detectors, offering a principled, data-driven alternative to conventional uncertainty modeling approaches reliant on fixed priors or Gaussian assumptions.
\end{enumerate}

\section{Related Work}
\subsection{Tiny Object Detection}
Advances in deep convolutional neural networks (DNNs) have significantly enhanced object detection tasks~\cite{lin2014microsoft}.
Despite the advances, tiny object detection remains a challenging problem due to the intrinsic limited pixel input~\cite{wang2021tiny}. The main difficulties include weak feature representation~\cite{lin2017feature}, information loss during downsampling~\cite{kisantal2019augmentation}, and a lower number of positive sample assignments resulting from increased sensitivity in IoU calculations~\cite{xu2022rfla}. Existing methods to address these issues can be broadly grouped into four categories: feature enhancement, data augmentation, scale-aware training, and super-resolution-based approaches.

\noindent{\bf Feature Enhancement.}
A major research direction focuses on improving multi-scale feature representation. SSD~\cite{liu2016ssd} detects objects using features at different resolutions. FPN~\cite{lin2017feature} introduces a top-down pathway with lateral connections to fuse semantic and spatial information across scales. This framework has been extended by methods like PANet~\cite{liu2018path} and Recursive-FPN~\cite{qiao2021detectors}. TridentNet~\cite{li2019scale} further enhances multi-scale detection by employing multiple branches with different receptive fields tailored to different object sizes. SET~\cite{sun2025set} amplifies the frequency signatures of tiny objects in a heterogeneous architecture.

\noindent{\bf Data Augmentation.}
Beyond standard augmentations (e.g., flipping, rotation, resizing), Kisantal \textit{et al.}~\cite{kisantal2019augmentation} improve detection by oversampling and copy-pasting tiny objects within training images. Recent developments in few-shot object detection (FSOD) also highlight the role of cross-modal knowledge transfer to alleviate data sparsity challenges in tiny object categories.

\noindent{\bf Scale-Aware Training.}
Detectors often struggle to maintain accuracy across object scales. SNIP~\cite{singh2018analysis} addresses this by restricting training to objects within specific scale intervals. UGS~\cite{sun2025uncertaintyawaregradientstabilizationsmall} reformulates object localization as a classification task to stabilize small objects' gradients.

\noindent{\bf Super-Resolution-Based Methods.}
Some methods enhance tiny object features through super-resolution techniques. PGAN~\cite{li2017perceptual} integrates GAN-based super-resolution into the detection pipeline. However, these approaches often incur high training and inference costs.
Recent strategies emphasize improved label assignment and proposal refinement~\cite{xu2022rfla}, which are critical for boosting recall and localization precision for tiny objects.

Orthogonal to existing TOD methods, our approach introduces a new perspective for tiny object detection by addressing annotation noise overfitting through flow-based uncertainty modeling.

\subsection{Learning with Noisy Labels}
Noise has emerged as a critical component in modern machine learning paradigms. From dropout layers injecting structural stochasticity to adversarial training harnessing perturbations for robustness, noise-driven mechanisms are now central to improving generalization, stability, and convergence in deep neural networks (DNNs)~\cite{goodfellow2016deep,PN,VPN,PiNGDA,NFIG,MiN}. 
Recent work further highlights how uncertainty, a form of implicit noise in predictions, can guide learning by exposing model weaknesses~\cite{kendall2017uncertainties}. 
These advances align with a shift toward beneficial noise learning, where controlled noise injection or exploitation enhances model performance. 
%
%

\noindent{\bf{Noise in Medical Learning.}}  Label noise poses a significant challenge in tasks like medical diagnosis, where inconsistent annotations can mislead models. Methods like DAL~\cite{li2023dynamics} introduce dynamics-aware loss functions that adaptively balance fitting ability and robustness, while self-paced learning frameworks~\cite{long2024interpretable} leverage medical guidelines to detect and mitigate label noise, improving interpretability and performance in multi-disease diagnosis tasks. These approaches highlight the importance of noise-aware learning in improving robustness and reliability in label-sensitive applications. 

\noindent\textbf{Noise in Multi-Modal Learning.}
Multi-modal learning faces significant challenges when dealing with noisy or incomplete data streams. Traditional approaches, such as incomplete multi-modal frameworks~\cite{kanwal2024incomplete}, address low-quality or missing signals by selectively downweighting unreliable channels while maintaining latent cross-modal consistency. 
Beyond these defensive strategies, recent research has revealed that carefully designed noise can actively enhance multi-modal learning. In vision-language models, deliberately injected noise strengthens cross-modal alignment by forcing robustness to perturbations~\cite{PiNI}. Similarly, in contrastive learning frameworks, common data augmentations can be reinterpreted as positive-incentive noise that improves representation learning~\cite{PiNDA}.
These approaches collectively demonstrate that noisy or missing modalities need not be treated as purely detrimental. Instead, by reframing such imperfections as opportunities for beneficial noise injection, multi-modal frameworks can achieve enhanced robustness, improved alignment, and superior generalization capabilities.

\noindent{\bf{Noise in Object Detection.}}  
Compared to image classification, object detection faces more diverse and complex label noise. This noise manifests primarily as four types: missing labels, extra labels, class shifts, and inaccurate bounding boxes. Some previous studies~\citep{chadwick2019training, li2020towards} assume all types of noise occur and try to tackle all types of noise simultaneously, while others~\citep{SDlocnet, OAMIL} focus on handling a specific type of noise (\textit{e.g.} inaccurate bounding box)~\citep{OAMIL}.

Collectively, these advances establish noise-robust learning as essential for safety critical noisy real-world settings. This imperative is especially critical for tiny object detection (TOD), where annotation noise compromises localization accuracy and stability, motivating our investigation.

\begin{figure}[t]
    \centering
    \includegraphics[width=1.0\linewidth]{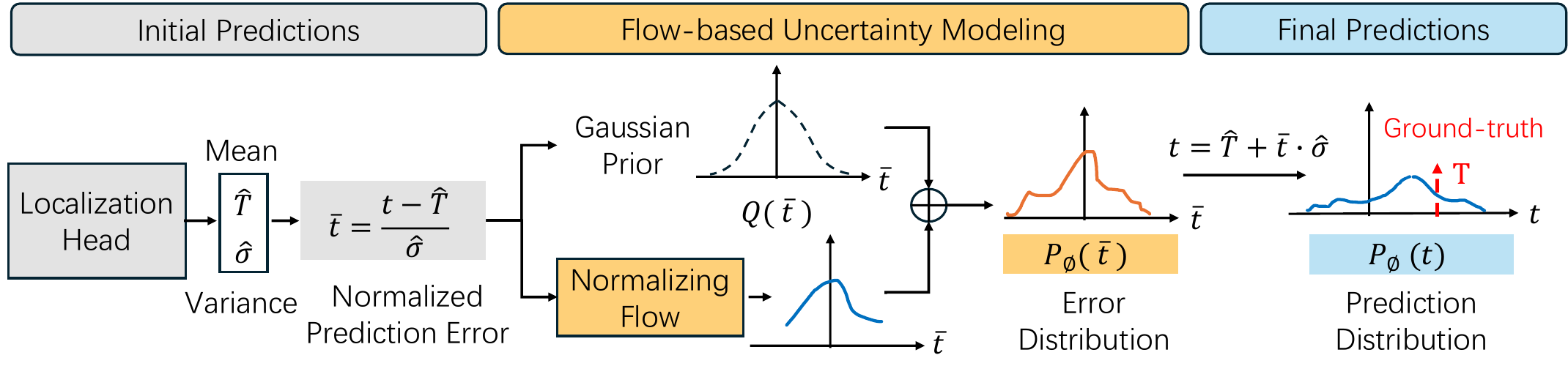}
    \caption{Overview of the noise-robust localization framework TOLF. The localization head predicts the mean $\hat{T}$ and uncertainty $\hat{\sigma}$ for each bounding box. The normalized prediction error is then modeled by a normalizing flow $G_{\phi}$ to capture complex, non-Gaussian error distributions. This enables robust estimation of the prediction distribution $P_\phi(t)$, which accounts for uncertainty and label noise.
    }
    \label{fig:framework}
\end{figure}

\section{Method}
\label{sec:method}
This section presents our approach to robust tiny object localization under noisy annotations. 
We first analyze limitations of existing localization uncertainty modeling paradigms (Sec.~\ref{subsec:preliminaries}), then introduce the \textbf{Tiny Object Localization Flow (TOLF)} framework that jointly learns flexible prediction distributions and uncertainty-aware optimization (Sec.~\ref{subsec:tolf}).
\subsection{Localization Uncertainty Modeling}
\label{subsec:preliminaries}
\noindent{\bf Conventional Localization.}  
Following previous detectors~\cite{girshick2014rich,girshick2015fast,ren2015faster}, we denote the localization targets and predictions as:  
\begin{equation}
\begin{split}
    \{T_{x}, T_{y}, T_{w}, T_{h}\} &= \{\frac{x - x_a}{w_a}, \frac{y - y_a}{h_a}, \log\frac{w}{w_a}, \log\frac{h}{h_a}\}, \\
    \{\hat{T}_{x}, \hat{T}_{y}, \hat{T}_{w}, \hat{T}_{h}\} &= \{\frac{\hat{x} - x_a}{w_a}, \frac{\hat{y} - y_a}{h_a}, \log\frac{\hat{w}}{w_a}, \log\frac{\hat{h}}{h_a}\},
\end{split}
\end{equation}
where \((x_a, y_a, w_a, h_a)\) denote the anchor coordinates, \((x, y, w, h)\) the ground-truth coordinates, and \((\hat{x}, \hat{y}, \hat{w}, \hat{h})\) the predicted coordinates, respectively. 
The $\mathcal{L}_2$ loss for $x$ can be formulated as:
\begin{equation}
    \begin{split}
        \mathcal{L}_{2}(T_{x}, \hat{T}_{x}) = \|T_{x} - \hat{T}_{x}\|_2^2,
    \end{split}
\label{eq1}
\end{equation}
which also can be applied to \(y\), \(w\), and \(h\).
To simplify, we use $T$ to denote the transformation parameters: $T = (T_x, T_y, T_w, T_h)$.
From a maximum likelihood estimation (MLE) perspective, the $\mathcal{L}_2$ loss assumes that the localization errors follow a homoscedastic Gaussian distribution:
\begin{equation}
P(T \mid \hat{T}) = \mathcal{N}(T; \hat{T}, \sigma^2),
\end{equation}
where the variance $\sigma^2$ is fixed and shared across all training samples. 
However, object localization often exhibits heteroscedasticity~\cite{li2022generalized}, where the localization uncertainty varies significantly across samples due to factors like object size, occlusion, or blur. 

\begin{figure*}[t]
    \centering
    \includegraphics[width=\linewidth]{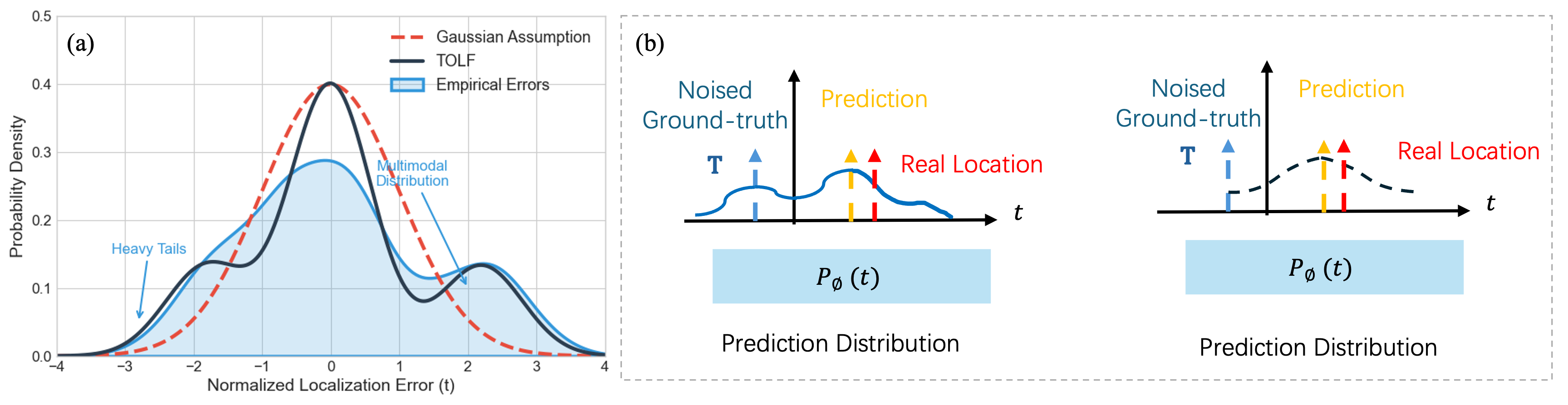}
    \caption{
        (a) Flexible distribution modeling enabled by TOLF, which better captures real-world noise compared to Gaussian assumptions. 
        (b) Illustration of noise-robust localization. The model predicts a distribution \( P_\theta(t) \) centered around the expected true location instead of regressing to noised Dirac ground-truths. This distributional supervision reduces overfitting to label noise and enabling uncertainty-aware localization.
    }
    \label{fig:reduce_overfit}
\end{figure*}
\noindent{\bf Gaussian-Based Modeling.}
Recent approaches address the localization uncertainty by explicitly modeling localization uncertainty.
To jointly learn localization and its confidence, \cite{he2019bounding} formulates bounding-box regression as minimizing the KL divergence $D_{\mathrm{KL}}(\cdot)$ between a Dirac ground-truth distribution $P_{D}$ and a predicted Gaussian distribution $P_{\Theta}$. The two distributions can be formulated as:
\begin{equation}
\label{eq:distributions}
P_{\Theta}(t) = \mathcal{N}\bigl(t \mid \hat{T},\sigma^{2}\bigr) = \frac{1}{\sqrt{2\pi}\sigma} \exp\left(-\frac{(t-\hat{T})^{2}}{2\sigma^{2}}\right), \quad P_{D}(t) = \delta(t - T),
\end{equation}
where $\hat{T}$ denotes the predicted mean, $T$ denotes the ground-truth localization target, and $\sigma$ the learned uncertainty.
Finally, the regression loss derives from the KL divergence:
\begin{align}
\label{eq:kl_loss}
\mathcal{L}_{\mathrm{reg}} &= D_{\mathrm{KL}}\bigl(P_{D},|,P_{\Theta}\bigr) = \frac{(T-\hat{T})^{2}}{2\sigma^{2}} + \frac{1}{2}\log\sigma^{2} + C,
\end{align}
with $C = \frac{1}{2}\log(2\pi)$ being a constant.
%
%
Note that the Gaussian form is fixed and symmetric, limiting its capacity to represent multi‐modal or long‐tailed annotation errors.   

\noindent{\bf Classification-Based Supervision.}  
Classification-based supervision can mitigate the impact of noisy labels, which can also be understood through the view of distribution. Generalized Focal Loss (GFL V1)~\cite{li2020generalized} introduces a classification paradigm for localization by quantizing continuous targets into discrete soft distributions. 
For a regression target $T \in [-\alpha, \alpha]$, the continuous range is partitioned into $n+1$ intervals with grid points $Y = \{\mathbf{y}_0, \mathbf{y}_1, \dots, \mathbf{y}_n\}$. 
Uncertainty is implicitly modeled through a categorical distribution generated by the localization head, where the network predicts logits $\mathbf{l} \in \mathbb{R}^{n+1}$, converted to probabilities via softmax:
\begin{equation}
    \mathbf{p}_i = \frac{\exp(\mathbf{l}_i)}{\sum_{k=0}^n \exp(\mathbf{l}_k)}, \quad \forall i \in \{0,...,n\}.
    \label{eq:softmax}
\end{equation}
Ground truth is encoded using two-hot targets based on adjacent grids $i_l$ and $i_r$:
\begin{equation}
    \mathbf{p}_i^* = 
    \begin{cases}
    |\mathbf{y}_i - T| \cdot \frac{n+1}{2\alpha}, & i = i_l, i_r \\
    0, & \text{otherwise},
    \end{cases}
    \label{eq:twohot}
\end{equation}
with optimization performed through cross-entropy:
\begin{equation}
    \mathcal{L}_\text{CE} = -\mathbf{p}_{i_l}^* \log \mathbf{p}_{i_l} - \mathbf{p}_{i_r}^* \log \mathbf{p}_{i_r}.
    \label{eq:gfl_loss}
\end{equation}
While modeling non-Gaussian uncertainty through discrete distributions, the two-hot target encoding (Eq.~\ref{eq:twohot}) imposes a piecewise linear structure that may not align with the true uncertainty in localization, limiting the flexibility and fidelity of the predicted distribution.
Also, the use of a fixed number of bins limits expressiveness in tails.

\subsection{TOLF: Tiny Object Localization Flow}
\label{subsec:tolf}
\noindent\textbf{Framework.}
To mitigate overfitting to noisy localization labels in tiny object detection, we propose TOLF, a flow-based framework that learns flexible prediction distributions with uncertainty estimation. The overview is shown in Fig.~\ref{fig:framework}.

TOLF introduces a probabilistic localization head outputs both predicted mean $\hat{T}_i$ and uncertainty $\hat{\sigma}_i$ for each bounding box coordinate. 
Following \cite{li2021human}, we model the distribution of normalized prediction error rather than predicted coordinates, which is defined as:
\begin{equation}  
    \bar{t}_i = \frac{T_i - \hat{T}_i}{\hat{\sigma}_i},  
\end{equation}  
where $T_i$ is the ground-truth value, $\hat{T}_i$ the predicted mean, and $\hat{\sigma}_i$ the predicted uncertainty.
To capture the complex characteristics of annotation noise, we model the distribution of $\bar{t}_i$ using normalizing flows. This framework transforms a simple base distribution (e.g., Gaussian) into a complex target distribution through a series of invertible mappings. 

Using normalizing flows, we model $\bar{t}_i$ with a flexible error distribution $P_\phi(\bar{t}_i)$. 
$P_\phi(\bar{t}_i)$ provides a flexible density approximator that overcomes parametric constraints through invertible transformations. 
Following \cite{li2021human}, we define the distribution as:
\begin{equation}
    P_\phi(\bar{t}_i) = Q(\bar{t}_i) \cdot G_\phi(\bar{t}_i) \cdot s,
\end{equation}
where $Q(\bar{t}_i) = \mathcal{N}(0,1)$ is a standard Gaussian prior, $G_\phi(\bar{t}_i)$ is the density correction learned by the flow model, and $s$ is a normalization constant to ensure $P_\phi$ integrates to one:
\begin{equation}
    s = \left( \int Q(\bar{t}) G_\phi(\bar{t}) d\bar{t} \right)^{-1}.
\end{equation}

Compared to conventional parametric models (e.g., Gaussian and Laplace), normalizing flows provide superior expressiveness, enabling modeling of capture skewness, heavy tails, and multi-modality in the distributions.
These properties are essential for robust localization under real-world annotation noise conditions.

The learned likelihood $P_\phi(\bar{t}_i)$ provides supervision through the negative log-likelihood loss. 
For each regression target $T_i$, the loss component is computed as:
\begin{equation}
\begin{aligned}
    \mathcal{L}_{\text{nf}}^{(i)} &= -\log P_{\Theta, \phi}(T_i | \mathcal{I}) \\
    &= -\log P_\phi(\bar{t}_i) + \log \hat{\sigma}_i \\
    &= -\log Q(\bar{t}_i) - \log G_\phi(\bar{t}_i) - \log s + \log \hat{\sigma}_i.
\end{aligned}
\end{equation}
The total regression loss is computed as the sum over all box parameters (e.g., $\{x, y, w, h\}$):
\begin{equation}
    \mathcal{L}_{\text{nf}} = \sum_{i \in \{x,y,w,h\}} \mathcal{L}_{\text{nf}}^{(i)}.
\end{equation}
This formulation enables the model to learn both mean and variance of regression targets while also allowing for flexible, non-Gaussian error modeling via the flow model.


\noindent\textbf{Uncertainty-Aware Gradient Modulation.}
TOLF facilitates noise robustness through dual gradient modulation.
\begin{align}
\frac{\partial \mathcal{L}_{\mathrm{nf}}^{(i)}}{\partial \hat{T}_i}
&= \frac{\partial \mathcal{L}_{\mathrm{nf}}^{(i)}}{\partial \bar{t}_i}
   \cdot \frac{\partial \bar{t}_i}{\partial \hat{T}_i}
\;=\;
\Bigl(-\partial_{\bar{t}_i}\log Q(\bar{t}_i)\;-\;\partial_{\bar{t}_i}\log G_\phi(\bar{t}_i)\Bigr)
\cdot\Bigl(-\tfrac{1}{\hat{\sigma}_i}\Bigr)
\nonumber\\
&= \frac{1}{\hat{\sigma}_i}
  \Bigl(\partial_{\bar{t}_i}\log Q(\bar{t}_i)\;+\;\partial_{\bar{t}_i}\log G_\phi(\bar{t}_i)\Bigr).
\label{eq:grad_mu}
\end{align}
The $1/\hat{\sigma}_i$ term adaptively attenuates gradients for high-uncertainty predictions, while $\partial\log G_\phi$ steers updates toward high-density regions of the learned error distribution. This suppresses updates for noisy labels while preserving stable updates for clean, well-localized objects. 
\begin{align}
\frac{\partial \mathcal{L}_{\mathrm{nf}}^{(i)}}{\partial \hat{\sigma}_i}
&= \frac{\partial \mathcal{L}_{\mathrm{nf}}^{(i)}}{\partial \bar{t}_i}
   \cdot \frac{\partial \bar{t}_i}{\partial \hat{\sigma}_i}
   \;+\;\frac{\partial}{\partial \hat{\sigma}_i}\bigl(\log \hat{\sigma}_i\bigr)
\nonumber\\
&= \Bigl(-\partial_{\bar{t}_i}\log Q(\bar{t}_i)\;-\;\partial_{\bar{t}_i}\log G_\phi(\bar{t}_i)\Bigr)
   \cdot\Bigl(-\tfrac{T_i - \hat{T}_i}{\hat{\sigma}_i^2}\Bigr)
   \;+\;\tfrac{1}{\hat{\sigma}_i}
\nonumber\\
&= \frac{T_i - \hat{T}_i}{\hat{\sigma}_i^2}
  \Bigl(\partial_{\bar{t}_i}\log Q(\bar{t}_i)\;+\;\partial_{\bar{t}_i}\log G_\phi(\bar{t}_i)\Bigr)
  \;+\;\tfrac{1}{\hat{\sigma}_i}.
\label{eq:grad_sigma}
\end{align}
Eqn.~\eqref{eq:grad_sigma} reveals TOLF's robust uncertainty learning. When large errors $|T_i-\hat{T}_i|$ originate from annotation noise rather than model error, the gradient term $\partial\log G_\phi$ reduces update magnitude, preventing excessive uncertainty inflation. 
Simultaneously, the $1/\hat{\sigma}_i$ component prevents uncertainty collapsing to zero, maintaining calibration.
Together, these two mechanisms form a balanced gradient modulation scheme that down‐weights noisy annotations while retaining stable updates for well‐localized  objects, thereby avoiding overfitting and ensuring reliable convergence.

\begin{table*}[!t]
\centering
\small
\caption{Main results with various frameworks on AI-TOD~\cite{wang2021tiny}. Models are trained on the AI-TOD \texttt{trainval} set and validated on the AI-TOD \texttt{test} set. We report APs (\%) under different IoU thresholds as well as APs (\%) for objects of various sizes based on the AI-TOD criterion. The \textbf{*} denotes using P2$\sim$P6 FPN features. The \textbf{bold} indicates the best result.}
\vspace{2mm}
\begin{tabularx}{\textwidth}{l|XXX|XXX}
\hline
Framework   & AP            & AP$_{0.5}$     & AP$_{0.75}$     & AP$_{ vt}$   & AP$_{ t}$     & AP$_{ s}$    \\ \hline
PAA~\cite{kim2020probabilistic}  & 10.0    & 26.5  & 6.7 & 3.5      & 10.5          & 13.1     \\
ATSS~\cite{zhang2020bridging} & 11.6    & 28.5 & 7.6             & 2.5 & 11.9          & 15.9     \\
Centernet~\cite{duan2019centernet} & 13.4          & 39.2   & 5.0      & 3.8          & 12.1          & 17.7               \\
DetectoRS~\cite{qiao2021detectors} & 14.8 & 32.8 & 11.5 & 0.0 & 10.8 & 18.3 \\
DotD~\cite{xu2021dot} & 16.1 &  39.2 & 10.6 & 8.3 & 17.6 & 18.1 \\
NWD~\cite{wang2021normalized} & 20.8 & 49.3 & 14.3 & 6.4  & 19.7 & 29.6        \\ 
SR-TOD~\cite{cao2024visible}  & 24.0 & 54.6 & 17.1 & 10.1 & 24.8 & 29.3 \\ \hline 
\multicolumn{7}{c}{One-stage} \\ \hline
FCOS~\cite{tian2019fcos}  & 12.0          & 29.0    & 8.0          & 2.5          & 11.9          & 17.1    \\
\textbf{w/ TOLF}  & \textbf{13.2} & \textbf{32.1} & \textbf{9.0}        & \textbf{3.2} & \textbf{13.5} & \textbf{19.4}    \\ \hline
FCOS*~\cite{tian2019fcos}   & 15.1   & 35.8  & 10.2    & 5.9          & 16.6          & 18.8       \\
\textbf{w/ TOLF}  & \textbf{16.7}   & \textbf{37.5} & \textbf{12.2} & \textbf{6.8} & \textbf{18.0} & \textbf{21.9}  \\ \hline
\multicolumn{7}{c}{Multi-stage} \\ \hline
Faster R-CNN~\cite{ren2015faster} & 11.1   & 26.3   & 8.1    & 0.0          & 7.2           & 23.3        \\ 
\textbf{w/ TOLF} & \textbf{12.8}  & \textbf{28.9}   & \textbf{10.3}       & \textbf{0.2} & \textbf{9.5} & \textbf{25.1}  \\ \hline 
Cascade R-CNN~\cite{cai2018cascade}  & 13.6          & 30.3   & 10.6            & 0.0          & 9.9          & 25.5      \\
\textbf{w/ TOLF} & \textbf{15.4}  & \textbf{33.1}   & \textbf{11.8}          & \textbf{0.5} & \textbf{11.3} & \textbf{27.5}    \\ \hline
RFLA~\cite{xu2022rfla}  & 21.7 & 50.5 & 15.3 & 8.3 & 21.8 & 24.5  \\
\textbf{w/ TOLF} &  \textbf{23.0}    & \textbf{53.2}      & \textbf{17.6}        & \textbf{10.1} & \textbf{23.7} & \textbf{27.9}   \\ \hline
\multicolumn{7}{c}{Transformer-based} \\ \hline
DINO-5scale~\cite{zhang2022dino} & 23.2 & 56.6 & 15.4 & 9.9 & 23.1 & 29.3 \\ 
\textbf{w/ TOLF} & \textbf{24.4} & \textbf{57.0} & \textbf{17.2} & \textbf{11.0} & \textbf{24.4} & \textbf{31.0} \\ 
\hline
\end{tabularx}
\label{ai-tod}
\end{table*}
\section{Experiments}
\subsection{Datasets and Implementation Details}
\noindent\textbf{Datasets}. 
We evaluate our method on three benchmark datasets: \textbf{AI-TOD}~\cite{wang2021tiny}, \textbf{DOTA-v2.0}~\cite{xia2018dota}, and \textbf{Tinyperson}~\cite{yu2020scale}. 
Our primary experiments are based on AI-TOD, a challenging dataset characterized by an average object size of only 12.8 pixels—significantly smaller than in standard detection datasets such as MS COCO (99.5 pixels)~\cite{lin2014microsoft}.
We also apply our method to DOTA-v2.0 and Tinyperson, both of which contain high-resolution aerial or drone imagery with a high density of tiny targets.

\noindent\textbf{Implementation Details}. We conducted the experiments on a computer with an NVIDIA RTX 3090 GPU. All CNN-based models utilize the ResNet-50~\cite{he2016deep} backbone, trained using the Stochastic Gradient Descent (SGD) optimizer for 12 epochs with 0.9 momentum, 0.0001 weight decay, and a batch size 2. The initial learning rate is 0.005, decaying at the 8th and 11th epochs. The data processing adheres to the default configurations of each dataset (e.g, fixed at 800$\times$800 for AI-TOD). 
We also train a transformer-based detector, DINO~\cite{zhang2022dino}, with 5-scale feature maps for 36 epochs as a baseline. The training uses an Adam optimizer with a weight decay of 0.0001, following the random crop and scale augmentation strategies of DETR~\cite{carion2020end}.

Our proposed localization paradigm is agnostic to the specific design of the normalizing flow.
In our experiments, we adopt RealNVP~\cite{dinh2017density} due to its fast and stable training behavior. 
We denote the invertible transformation as a fully-connected architecture with $L_\text{fc}$ layers and $N_\text{n}$ neurons per layer, i.e., $L_\text{fc} \times N_\text{n}$. 
By default, we set $L_\text{fc} = 3$ and $N_\text{n} = 64$. 
This flow model is lightweight and introduces negligible overhead to the overall training process.

\begin{table*}[t!]
\centering
\small
\caption{Detection results with various frameworks on TinyPerson~\cite{yu2020scale}. All models are trained on the {\tt train} set and evaluated on the {\tt val} set. We report AP (\%) at different IoU thresholds and across object sizes following the TinyPerson benchmark. \textbf{Bold} denotes the best result within each base detector group.}
\vspace{2mm}
\begin{tabularx}{\textwidth}{l|XXXXXXX}
\hline
Framework & AP$_{50}^{tiny}$   & AP$_{50}^{tiny1}$ & AP$_{50}^{tiny2}$ & AP$_{50}^{tiny3}$ & AP$_{50}^{small}$  & AP$_{25}^{tiny}$  & AP$_{75}^{tiny}$  \\ \hline
FCOS~\cite{tian2019fcos}         & 16.9 & 3.9   & 12.4   & 29.3 & 36.8 & 40.5 & 1.5  \\
Faster R-CNN~\cite{ren2015faster} & 43.6 & 48.3 & 53.5 & 43.6 & 56.7 & 64.1 & 5.4  \\ \hline
RetinaNet~\cite{lin2017focal}    & 15.5 & 3.0   & 14.4   & 29.1 & 46.8 & 48.4 & 1.3  \\
\textbf{RetinaNet w/ TOLF}       & \textbf{17.2}  & \textbf{3.7}  & \textbf{15.8}  & \textbf{29.6}  & \textbf{47.3}  & \textbf{51.6}  & \textbf{1.5}  \\ \hline
AutoAssign~\cite{zhu2020autoassign} & 21.0 & 7.1   & 19.7   & 32.3 & 48.1 & 55.0 & 1.4  \\
\textbf{AutoAssign w/ TOLF}      & \textbf{22.1}  & \textbf{7.5}  & \textbf{20.8} & \textbf{33.2}  & \textbf{50.3} & \textbf{57.3}  & \textbf{2.0}  \\ \hline
\end{tabularx}
\label{tinyperson}
\end{table*}

\subsection{Results on AI-TOD}
We evaluate TOLF across multiple detectors on the AI-TOD benchmark~\cite{xu2022detecting}, comparing against state-of-the-art TOD methods.
As shown in Tab.~\ref{ai-tod}, TOLF consistently improves all baselines by $\sim$2\% AP.
Notably, it enhances the one-stage FCOS~\cite{tian2019fcos} detector by 1.2\% AP and 1.6\% AP$_{t}$.
When incorporating P2$\sim$P6 FPN features—a representative TOD configuration leveraging high-resolution P2 for tiny objects—TOLF further boosts FCOS by 1.6\% AP.
TOLF also generalizes effectively to multi-stage detectors, improving Faster R-CNN~\cite{ren2015faster} and Cascade R-CNN~\cite{cai2018cascade} by 1.7\% AP and 1.8\% AP, respectively.
Critically, TOLF complements the state-of-the-art RFLA~\cite{xu2022rfla} method, adding a 1.3\% AP gain.
For transformer-based detectors, TOLF achieves 24.4\% AP on DINO-5scale~\cite{zhang2022dino} (a 1.2\% AP increase), outperforming competitors including DotD~\cite{xu2021dot}, NWD~\cite{wang2021normalized}, and SR-TOD~\cite{cao2024visible}.

\begin{table}[h]
\centering
\small
\caption{Detection performance on DOTA-v2.0~\cite{xia2018dota}. All models are trained on the DOTA-v2.0 {\tt train} set and evaluated on the {\tt val} set. TOLF is applied to four representative base detectors. \textbf{Bold} indicates the best result for each group.}
\vspace{2mm}
\begin{tabularx}{\textwidth}{l|X|XXX}
\hline
Framework  & AP  & AP$_{ vt}$ & AP$_{ t}$  & AP$_{ s}$  \\ \hline
ATSS~\cite{zhang2020bridging}    & 32.7  & 0.7 & 6.9 & 23.4   \\
\textbf{ATSS w/ TOLF} & \textbf{34.1} & \textbf{0.8} & \textbf{7.3} & \textbf{24.3} \\ \hline
Faster R-CNN~\cite{ren2015faster}  & 35.6   & 0.0 & 7.1 & 28.9 \\
\textbf{Faster R-CNN w/ TOLF} & \textbf{36.5} & \textbf{0.4} & \textbf{7.5} & \textbf{29.5} \\ \hline
FCOS~\cite{tian2019fcos}  & 31.8 & 0.3 & 4.0 & 19.4  \\
FCOS w/ RFLA~\cite{xu2022rfla}   & 32.1 & \textbf{0.7} & 6.8 & 23.6  \\
\textbf{FCOS w/ TOLF} & \textbf{33.3} & 0.6 & \textbf{7.1} & \textbf{24.8} \\ \hline
AutoAssign~\cite{zhu2020autoassign}   & 33.8  & 0.9 & 7.3 & 22.4  \\
\textbf{AutoAssign w/ TOLF} & \textbf{35.5} & \textbf{1.2} & \textbf{8.7} & \textbf{23.9}  \\ 
\hline
\end{tabularx}
\label{dota}
\end{table}

\subsection{Results on DOTA-v2.0 and TinyPerson}
We evaluate the effectiveness of TOLF on two challenging TOD benchmarks: DOTA-v2.0~\cite{xia2018dota} and TinyPerson~\cite{yu2020scale}, both of which feature densely packed, low-resolution objects.
As shown in Tab.~\ref{dota}, TOLF consistently improves multiple detectors on DOTA-v2.0.
With FCOS, TOLF yields a 1.5\% AP improvement, including 0.6\% in AP$_{vt}$ and 0.7\% in AP$_t$.
On top of AutoAssign, TOLF provides an even larger boost of 1.7\% AP, with 0.3\% and 1.4\% gains in AP$_{vt}$ and AP$_t$, respectively.
Compared to the prior art RFLA~\cite{xu2022rfla} and DCFL~\cite{xu2023dynamic}, TOLF outperforms both, achieving 3.2\% and 1.5\% higher AP than RFLA and DCFL when used with FCOS.

Tab.~\ref{tinyperson} reports results on the TinyPerson dataset.
TOLF brings consistent gains across two diverse baselines.
With RetinaNet, TOLF improves AP$_{50}^{tiny}$ by 1.7\% and increases AP$_{25}^{tiny}$ by 3.2\%.
When applied to AutoAssign, TOLF achieves a substantial 1.1\% AP$_{50}^{tiny}$ improvement and boosts performance across all subcategories.
These results highlight TOLF's generalizability and robustness across architectures and real-world scenarios.

\begin{table}[h]
\centering
\small
\caption{Detection performance on MS COCO~\cite{lin2014microsoft}. Note that models are trained on COCO {\tt train2017} and validated on COCO {\tt val2017}.}
\vspace{2mm}
\begin{tabular}{l|l|llll}
\hline
Framework  & AP & AP$_{vt}$  & AP$_{t}$ & AP$_{s}$ & AP$_{m}$ \\ \hline
FCOS~\cite{tian2019fcos} & 36.4 & 7.9 & 19.6 & 27.2 & 43.6  \\
FCOS w/ TOLF & 37.2 & 8.9 & 20.8 & 28.3 & 44.5  \\
\hline
\end{tabular}
\label{coco}
\end{table}

\subsection{Results on COCO}
We further verify TOLF's performance on MS COCO~\cite{lin2014microsoft}, a large-scale benchmark. As shown in Tab.~\ref{coco}, TOLF brings substantial improvements over the FCOS baseline, achieving +0.8 AP overall. The enhancements are particularly significant for tiny objects under AI-TOD metrics, with +1.0 AP\textsubscript{vt} for very tiny objects and +1.2 AP\textsubscript{t} for tiny objects. These results confirm TOLF's effectiveness not only on tiny object datasets but also on general object detection benchmarks.

\begin{figure}[t]
    \centering
    \includegraphics[width=1.0\linewidth]{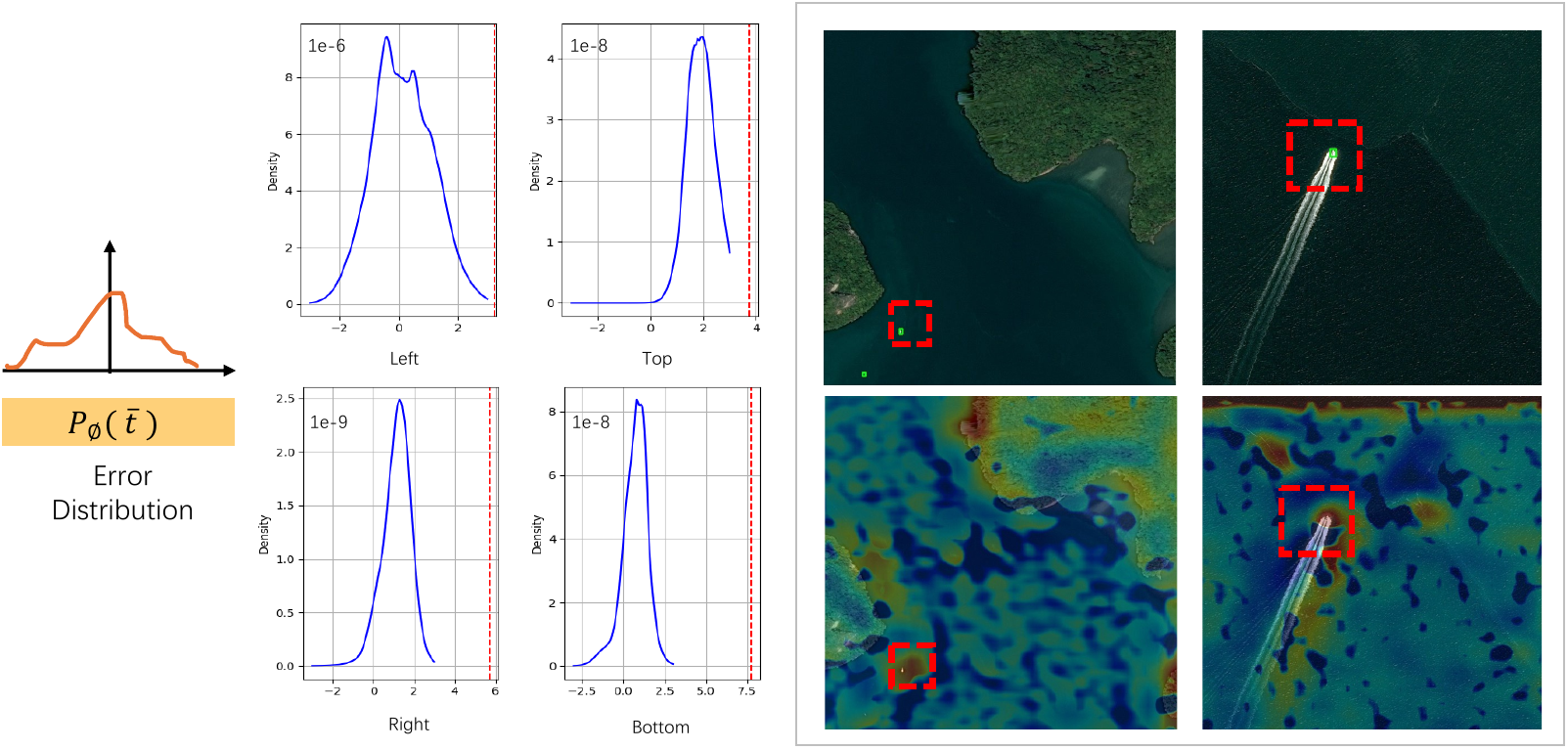}
    \caption{
        \textbf{Left.}
       Learned 1D error distributions for bounding box coordinates (left, right, top, bottom) using normalizing flows. Each distribution \( P_{\theta}(\overline{t}) \) plots the residual error for a specific coordinate (e.g., left boundary), conditioned on fixed values of the other coordinates. 
       %
        %
        %
        \textbf{Right.}
        Average predicted variance \( \sigma \) of four coordinates overlaid with input.
    }
    \label{fig:predicted_errors_sigmas}
\end{figure}
\begin{table}[t!]
\centering
\small
\caption{Ablation study of TOLF components on AI-TOD test set. All experiments use FCOS* as baseline.}
\vspace{2mm}
\begin{tabular}{l|c|c|c|c|c|c}
\hline
\textbf{Variant ($\lambda$)} & \textbf{AP} & \textbf{AP$_{0.5}$} & \textbf{AP$_{0.75}$} & \textbf{AP$_{vt}$} & \textbf{AP$_{t}$} & \textbf{AP$_{s}$} \\ 
\hline
FCOS* & 15.1 & 35.8 & 10.2 & 5.9 & 16.6 & 18.8 \\
\hline
\multicolumn{7}{c}{\textit{Core Components}} \\ 
\hline
+ Normalizing Flow  & 15.9 & 36.7 & 10.8 & 6.3 & 17.0 & 19.7 \\
+ Uncertainty & 15.7 & 36.2 & 10.6 & 6.1 & 16.8 & 19.3 \\
+ TOLF ($\lambda=0.1$) & \textbf{16.7} & \textbf{37.5} & \textbf{12.2} & \textbf{6.8} & \textbf{18.0} & \textbf{21.9} \\
\hline
\multicolumn{7}{c}{\textit{Localization Uncertainty Comparison}} \\ 
\hline
w/ KL Loss~\cite{he2019bounding}& 15.6 & 36.2 & 10.9 & 5.7 & 17.2 & 20.5 \\
w/ GFocal~\cite{li2022generalized} & 16.0 & 36.8 & 11.4 & 6.0 & 17.6 & 20.7 \\
\hline
\multicolumn{7}{c}{\textit{Loss Weight}} \\ 
\hline
TOLF ($\lambda=0.01$) & 16.1 & 36.7 & 11.6 & 6.3 & 17.3 & 20.9 \\
TOLF ($\lambda=0.1$)  & \textbf{16.7} & \textbf{37.5} & \textbf{12.2} & \textbf{6.8} & \textbf{18.0} & \textbf{21.9} \\
TOLF ($\lambda=1.0$)  & 16.2 & 36.6 & 11.0 & 6.0 & 17.4 & 20.6 \\
\hline
\end{tabular}
\label{tab:ablation}
\end{table}

As shown in Fig.~\ref{fig:predicted_errors_sigmas}, the learned per-coordinate residual distributions are non-Gaussian, asymmetric. For the left coordinate, two modes appear near $-0.9$ and $0.9$. High density near $0$ indicates consistent annotations, whereas heavy tails indicate ambiguity or outliers. These results demonstrate that the normalizing-flow model captures annotation noise beyond Gaussian assumptions.

\vspace{5mm}
\subsection{Visualizations}
As shown in Fig.~\ref{fig:predicted_errors_sigmas}, the learned per-coordinate residual distributions are non-Gaussian and asymmetric. For the left coordinate, two peaks appear near $-0.9$ and $0.9$. High density near $0$ indicates consistent annotations, whereas heavy tails indicate ambiguity or outliers. These results demonstrate that the normalizing flow model captures annotation noise beyond Gaussian assumptions.
The red boxed area highlights regions of elevated variance, indicating localized uncertainty that modulates distribution sharpness. This guides noise-aware training through gradient attenuation in high-uncertainty regions, enhancing robustness against noised annotations.

Fig.~\ref{fig:reduce_overfit} demonstrates significant improvements from the TOLF design in challenging scenarios. 
Compared to the baseline, our approach achieves more accurate localization and reduces false positives in cluttered environments, confirming TOLF's effectiveness in enhancing detection reliability under real-world noise conditions.
\begin{table*}[h]
\centering
\caption{Efficiency analysis of TOLF components on AI-TOD dataset.}
\vspace{2mm}
\setlength{\tabcolsep}{4pt}
\begin{tabular}{@{}l ccc ccc@{}}
\toprule
\textbf{Method} & \multicolumn{3}{c}{\textbf{Performance} (\%)} & \multicolumn{3}{c}{\textbf{Cost}} \\
\cmidrule(lr){2-4} \cmidrule(l){5-7}
 & AP & AP\textsubscript{vt} & AP\textsubscript{t} & Time (ms) & GFLOPs & Params (M) \\
\midrule
FCOS Baseline & 12.0 & 2.5 & 11.9 & 17.2 & 126.1 & 37.0 \\
+ TOLF (full) & 13.7 & 2.9 & 13.1 & 19.8 & 127.4 & 37.5 \\
+ TOLF (simplified) & 13.5 & 2.8 & 13.0 & 18.1 & 126.8 & 37.2 \\
\bottomrule
\end{tabular}
\label{tab:efficiency}
\end{table*}
\subsection{Ablation Study}
To evaluate the effectiveness of each component in our framework, we conduct extensive ablation studies on the AI-TOD test set using FCOS* as the baseline. Results are reported in Tab.~\ref{tab:ablation}.

\noindent{\bf Component Analysis.}   We first evaluate the individual effects of TOLF’s core components.
Applying the normalizing flow to model residual errors using a log-likelihood objective ($\mathcal{L}_{\text{flow}} = -\log P_\phi(T_i - \hat{T}_i)$) improves AP from 15.1\% to 15.9\%, demonstrating that flexible, data-driven error modeling beyond fixed Gaussian assumptions enhances robustness under annotation noise.
Incorporating uncertainty-aware weighting into the final loss function ($\mathcal{L}_{\text{uncertainty}} = |T_i - \hat{T}_i| / \sigma_i$) yields a comparable improvement to 15.7\%, suggesting that adaptively modulating gradients based on predicted uncertainty mitigates the influence of noisy or ambiguous labels.
Combining both components leads to the full TOLF framework, achieving 16.7\% AP—an absolute improvement of +1.6\% over the baseline with consistent gains across AP$_{0.5}$, AP$_{0.75}$, and all object scales (vt/t/s).

\noindent{\bf Comparison with Other Losses.} We compare TOLF’s likelihood-based loss against popular uncertainty-aware losses.: KL Loss~\cite{he2019bounding} and GFocal~\cite{li2022generalized}.
While both alternatives outperform the baseline, TOLF achieves higher accuracy across all metrics, showing that explicitly modeling residual error distributions with flows leads to more robust localization than assuming predefined error forms.

\noindent{\bf Effect of TOLF’s loss weight $\lambda$.} We further study the effect of the uncertainty modulation weight $\lambda$.
As shown in Tab.~\ref{tab:ablation}, a small value $\lambda=0.1$ performs best, striking a balance between preserving useful gradients and suppressing noise-prone updates.

\noindent\textbf{Efficiency Analysis.} 
We show that the proposed TOLF framework introduces minimal computational overhead while achieving performance improvements. 
As shown in Tab.~\ref{tab:efficiency}, the full TOLF implementation increases inference time by only 15.1\% (+2.6 ms) and computational complexity by 1.0\%, while improving AP by 1.7\%.  
To further optimize efficiency, we explore a simplified RealNVP~\cite{dinh2017density} configuration that reduces the number of coupling layers from 6 to 3 and employs shallower neural networks within each transformation block.
This simplified variant maintains 98.5\% of the performance gain while reducing the additional latency to just 5.2\%. 
The marginal performance trade-off demonstrates the potential for deploying TOLF in resource-constrained environments without compromising its core effectiveness.

\begin{table}[t!]
\centering
\caption{Performance comparison under different patch sizes on AI-TOD dataset}
\vspace{2mm}
\label{tab:patch_size_ablation}
\begin{tabular}{l|l|cccc}
\hline
\textbf{Patch Size} & \textbf{Method} & \textbf{AP} & \textbf{AP\textsubscript{vt}} & \textbf{AP\textsubscript{t}} & \textbf{AP\textsubscript{s}} \\
\hline
\multirow{2}{*}{8×8} & DINO-5scale & 22.8 & 9.2 & 22.7 & 29.0 \\
& TOLF & \textbf{24.2} & \textbf{10.3} & \textbf{24.1} & \textbf{30.5} \\
\hline
\multirow{2}{*}{16×16} & DINO-5scale & 22.3 & 8.7 & 22.2 & 28.5 \\
& TOLF & \textbf{24.4} & \textbf{10.5} & \textbf{24.3} & \textbf{30.7} \\
\hline
\multirow{2}{*}{32×32} & DINO-5scale & 20.1 & 6.5 & 20.0 & 26.8 \\
& TOLF & \textbf{23.0} & \textbf{9.1} & \textbf{22.9} & \textbf{29.4} \\
\hline
\end{tabular}
\label{tab:patch_size}
\end{table}

\noindent\textbf{Robustness to Patch Size Variations.}
Recent transformer-based networks have demonstrated sensitivity to patch size variations~\cite{chen2025hieratok}, which can impact feature granularity and capacity~\cite{shi20252d}. We investigate patch size variations in TOD and evaluate TOLF under identical settings to assess robustness.
As shown in Tab.~\ref{tab:patch_size}, reducing the patch size to $8{\times}8$ yields marginal gains for DINO-5scale, primarily benefiting tiny objects. 
%
%
TOLF improves consistently across all configurations and remains stable at $32{\times}32$, outperforming DINO-5scale by $2.9$~AP
This demonstrates TOLF's ability to preserve detection quality despite coarser feature representations.

\vspace{5mm}
\section{Conclusions}
In this paper, we address the challenge of robust tiny object localization under annotation noise, a critical yet underexplored issue.
We show that conventional tiny object detectors are highly sensitive to noisy labels, particularly when trained with strict localization objectives that inadvertently promote overfitting.
To tackle this, we introduce TOLF, a noise-robust localization framework that models residual errors with normalizing flows and suppresses unreliable supervision via uncertainty-guided optimization.
TOLF enables flexible, non-Gaussian error modeling through invertible transformations and incorporates uncertainty-aware gradient modulation to down-weight high-variance, noise-prone predictions.
Extensive experiments across three challenging benchmarks demonstrate that TOLF consistently improves detection accuracy for tiny objects.
This work highlights the importance of flexible label noise modeling for improving the reliability of tiny object detectors.

\bibliographystyle{elsarticle-num} 
\bibliography{main.bib}

@String(PAMI = {IEEE Trans. Pattern Anal. Mach. Intell.})

@String(CVPR= {IEEE Conf. Comput. Vis. Pattern Recog.})

@String(ICCV= {Int. Conf. Comput. Vis.})

@String(ECCV= {Eur. Conf. Comput. Vis.})

@String(NIPS= {Adv. Neural Inform. Process. Syst.})

@String(ICPR = {Int. Conf. Pattern Recog.})

@String(ICLR = {Int. Conf. Learn. Represent.})

@String(AAAI = {AAAI})

@String(WACV = {WACV})

@String(PAMI  = {IEEE TPAMI})

@String(CVPR  = {CVPR})

@String(ICCV  = {ICCV})

@String(ECCV  = {ECCV})

@String(NIPS  = {NeurIPS})

@String(ICPR  = {ICPR})

@String(ICLR  = {ICLR})

@String(ICML = {ICML})

@inproceedings{ren2015faster,
  title={Faster r-cnn: Towards real-time object detection with region proposal networks},
  author={Ren, Shaoqing and He, Kaiming and Girshick, Ross and Sun, Jian},
  booktitle=NIPS,
  volume={28},
  year={2015}
}

@inproceedings{lin2017feature,
  title={Feature pyramid networks for object detection},
  author={Lin, Tsung-Yi and Doll{\'a}r, Piotr and Girshick, Ross and He, Kaiming and Hariharan, Bharath and Belongie, Serge},
  booktitle=CVPR,
  pages={2117--2125},
  year={2017}
}

@inproceedings{lin2017focal,
  title={Focal loss for dense object detection},
  author={Lin, Tsung-Yi and Goyal, Priya and Girshick, Ross and He, Kaiming and Doll{\'a}r, Piotr},
  booktitle=ICCV,
  pages={2980--2988},
  year={2017}
}

@inproceedings{kisantal2019augmentation,
  title={Augmentation for small object detection},
  author={Kisantal, Mate and Wojna, Zbigniew and Murawski, Jakub and Naruniec, Jacek and Cho, Kyunghyun and others},
  booktitle={CS \& IT Conference Proceedings},
  volume={9},
  number={17},
  year={2019},
  organization={CS \& IT Conference Proceedings}
}

@inproceedings{li2019scale,
  title={Scale-aware trident networks for object detection},
  author={Li, Yanghao and Chen, Yuntao and Wang, Naiyan and Zhang, Zhaoxiang},
  booktitle=ICCV,
  pages={6054--6063},
  year={2019}
}

@inproceedings{singh2018analysis,
  title={An analysis of scale invariance in object detection snip},
  author={Singh, Bharat and Davis, Larry S},
  booktitle=CVPR,
  pages={3578--3587},
  year={2018}
}

@inproceedings{xu2021dot,
  title={Dot Distance for Tiny Object Detection in Aerial Images},
  author={Xu, Chang and Wang, Jinwang and Yang, Wen and Yu, Lei},
  booktitle=CVPR,
  pages={1192--1201},
  year={2021}
}

@article{wang2021normalized,
  title={A normalized Gaussian Wasserstein distance for tiny object detection},
  author={Wang, Jinwang and Xu, Chang and Yang, Wen and Yu, Lei},
  journal={arXiv preprint arXiv:2110.13389},
  year={2021}
}

@inproceedings{cai2018cascade,
  title={Cascade r-cnn: Delving into high quality object detection},
  author={Cai, Zhaowei and Vasconcelos, Nuno},
  booktitle=CVPR,
  pages={6154--6162},
  year={2018}
}

@inproceedings{lin2014microsoft,
  title={Microsoft coco: Common objects in context},
  author={Lin, Tsung-Yi and Maire, Michael and Belongie, Serge and Hays, James and Perona, Pietro and Ramanan, Deva and Doll{\'a}r, Piotr and Zitnick, C Lawrence},
  booktitle=ECCV,
  pages={740--755},
  year={2014},
  organization={Springer}
}

@inproceedings{he2015convolutional,
  title={Convolutional neural networks at constrained time cost},
  author={He, Kaiming and Sun, Jian},
  booktitle=CVPR,
  pages={5353--5360},
  year={2015}
}

@inproceedings{yang2022querydet,
  title={QueryDet: Cascaded sparse query for accelerating high-resolution small object detection},
  author={Yang, Chenhongyi and Huang, Zehao and Wang, Naiyan},
  booktitle=CVPR,
  pages={13668--13677},
  year={2022}
}

@inproceedings{girshick2015fast,
  title={Fast r-cnn},
  author={Girshick, Ross},
  booktitle=ICCV,
  pages={1440--1448},
  year={2015}
}

@inproceedings{girshick2014rich,
  title={Rich feature hierarchies for accurate object detection and semantic segmentation},
  author={Girshick, Ross and Donahue, Jeff and Darrell, Trevor and Malik, Jitendra},
  booktitle=CVPR,
  pages={580--587},
  year={2014}
}

@inproceedings{he2019bounding,
  title={Bounding box regression with uncertainty for accurate object detection},
  author={He, Yihui and Zhu, Chenchen and Wang, Jianren and Savvides, Marios and Zhang, Xiangyu},
  booktitle=CVPR,
  pages={2888--2897},
  year={2019}
}

@inproceedings{he2016deep,
  title={Deep residual learning for image recognition},
  author={He, Kaiming and Zhang, Xiangyu and Ren, Shaoqing and Sun, Jian},
  booktitle=CVPR,
  pages={770--778},
  year={2016}
}

@article{li2022generalized,
  title={Generalized Focal Loss: Towards Efficient Representation Learning for Dense Object Detection},
  author={Li, Xiang and Lv, Chengqi and Wang, Wenhai and Li, Gang and Yang, Lingfeng and Yang, Jian},
  journal=PAMI,
  year={2022},
  publisher={IEEE}
}

@inproceedings{zhang2020bridging,
  title={Bridging the gap between anchor-based and anchor-free detection via adaptive training sample selection},
  author={Zhang, Shifeng and Chi, Cheng and Yao, Yongqiang and Lei, Zhen and Li, Stan Z},
  booktitle=CVPR,
  pages={9759--9768},
  year={2020}
}

@inproceedings{kim2020probabilistic,
  title={Probabilistic anchor assignment with iou prediction for object detection},
  author={Kim, Kang and Lee, Hee Seok},
  booktitle=ECCV,
  pages={355--371},
  year={2020}
}

@inproceedings{duan2019centernet,
  title={Centernet: Keypoint triplets for object detection},
  author={Duan, Kaiwen and Bai, Song and Xie, Lingxi and Qi, Honggang and Huang, Qingming and Tian, Qi},
  booktitle=ICCV,
  pages={6569--6578},
  year={2019}
}

@inproceedings{carion2020end,
  title={End-to-end object detection with transformers},
  author={Carion, Nicolas and Massa, Francisco and Synnaeve, Gabriel and Usunier, Nicolas and Kirillov, Alexander and Zagoruyko, Sergey},
  booktitle=ECCV,
  pages={213--229},
  year={2020},
}

@inproceedings{li2020generalized,
  title={Generalized focal loss: Learning qualified and distributed bounding boxes for dense object detection},
  author={Li, Xiang and Wang, Wenhai and Wu, Lijun and Chen, Shuo and Hu, Xiaolin and Li, Jun and Tang, Jinhui and Yang, Jian},
  booktitle=NIPS,
  year={2020}
}

@inproceedings{du2023adaptive,
  title={Adaptive sparse convolutional networks with global context enhancement for faster object detection on drone images},
  author={Du, Bowei and Huang, Yecheng and Chen, Jiaxin and Huang, Di},
  booktitle=CVPR,
  pages={13435--13444},
  year={2023}
}

@inproceedings{tian2019fcos,
  title={Fcos: Fully convolutional one-stage object detection},
  author={Tian, Zhi and Shen, Chunhua and Chen, Hao and He, Tong},
  booktitle=ICCV,
  pages={9627--9636},
  year={2019}
}

@inproceedings{xu2022rfla,
  title={RFLA: Gaussian receptive field based label assignment for tiny object detection},
  author={Xu, Chang and Wang, Jinwang and Yang, Wen and Yu, Huai and Yu, Lei and Xia, Gui-Song},
  booktitle=ECCV,
  pages={526--543},
  year={2022},
}

@inproceedings{liu2016ssd,
	title={Ssd: Single shot multibox detector},
	author={Liu, Wei and Anguelov, Dragomir and Erhan, Dumitru and Szegedy, Christian and Reed, Scott and Fu, Cheng-Yang and Berg, Alexander C},
	booktitle=ECCV,
   pages={21--37},
	year={2016},
}

@article{PN,
  title={Positive-incentive noise},
  author={Li, Xuelong},
  journal={IEEE Transactions on Neural Networks and Learning Systems},
  volume={35},
  number={6},
  pages={8708--8714},
  year={2024},
  publisher={IEEE}
}

@article{VPN,
    title={Variational Positive-incentive Noise: How Noise Benefits Models}, 
    author={Hongyuan Zhang and Sida Huang and Yubin Guo and Xuelong Li},
    year={2025},
    journal={IEEE Transactions on Pattern Analysis and Machine Intelligence},
}

@article{PiNDA,
  title={Data augmentation of contrastive learning is estimating positive-incentive noise},
  author={Zhang, Hongyuan and Xu, Yanchen and Huang, Sida and Li, Xuelong},
  journal={arXiv preprint arXiv:2408.09929},
  year={2024}
}

@inproceedings{PiNI,
  title={Enhance vision-language alignment with noise},
  author={Huang, Sida and Zhang, Hongyuan and Li, Xuelong},
  booktitle={Proceedings of the AAAI Conference on Artificial Intelligence},
  volume={39},
  number={16},
  pages={17449--17457},
  year={2025}
}

@inproceedings{PiNGDA,
  title={Learn Beneficial Noise as Graph Augmentation},
  author={Huang, Siqi and Xu, Yanchen and Zhang, Hongyuan and Li, Xuelong},
  booktitle={Proceedings of the 42nd International Conference on Machine Learning (ICML)},
  year={2025},
  pages={},
}

@article{MiN,
  title={Mixture of Noise for Pre-Trained Model-Based Class-Incremental Learning}, 
  author={Kai Jiang and Zhengyan Shi and Dell Zhang and Hongyuan Zhang and Xuelong Li},
  journal={arXiv preprint   arXiv:2509.16738},
  year={2025},
}

@article{NFIG,
  title={Nfig: Autoregressive image generation with next-frequency prediction},
  author={Huang, Zhihao and Qiu, Xi and Ma, Yukuo and Zhou, Yifu and Chen, Junjie and Zhang, Hongyuan and Zhang, Chi and Li, Xuelong},
  journal={arXiv preprint arXiv:2503.07076},
  year={2025}
}

@inproceedings{liu2018path,
  title={Path aggregation network for instance segmentation},
  author={Liu, Shu and Qi, Lu and Qin, Haifang and Shi, Jianping and Jia, Jiaya},
  booktitle=CVPR,
  pages={8759--8768},
  year={2018}
}

@article{zhang2022dino,
  title={Dino: Detr with improved denoising anchor boxes for end-to-end object detection},
  author={Zhang, Hao and Li, Feng and Liu, Shilong and Zhang, Lei and Su, Hang and Zhu, Jun and Ni, Lionel M and Shum, Heung-Yeung},
  journal=ICLR,
  year={2023}
}

@misc{goodfellow2016deep,
  title={Deep learning},
  author={Goodfellow, Ian},
  year={2016},
  publisher={MIT press}
}

@article{kendall2017uncertainties,
  title={What uncertainties do we need in bayesian deep learning for computer vision?},
  author={Kendall, Alex and Gal, Yarin},
  journal={Advances in neural information processing systems},
  volume={30},
  year={2017}
}

@inproceedings{sun2025set,
  title={SET: Spectral Enhancement for Tiny Object Detection},
  author={Sun, Huixin and Wang, Runqi and Li, Yanjing and Yang, Linlin and Lin, Shaohui and Cao, Xianbin and Zhang, Baochang},
  booktitle={Proceedings of the Computer Vision and Pattern Recognition Conference},
  pages={4713--4723},
  year={2025}
}

@misc{sun2025uncertaintyawaregradientstabilizationsmall,
      title={Uncertainty-Aware Gradient Stabilization for Small Object Detection}, 
      author={Huixin Sun and Yanjing Li and Linlin Yang and Xianbin Cao and Baochang Zhang},
      year={2025},
      eprint={2303.01803},
      archivePrefix={arXiv},
      primaryClass={cs.CV},
      url={https://arxiv.org/abs/2303.01803}, 
}

@inproceedings{wang2021tiny,
  title={Tiny object detection in aerial images},
  author={Wang, Jinwang and Yang, Wen and Guo, Haowen and Zhang, Ruixiang and Xia, Gui-Song},
  booktitle=ICPR,
  pages={3791--3798},
  year={2021}}

@article{zhou2025spatial,
  title={Spatial residual for underwater object detection},
  author={Zhou, Jingchun and He, Zongxin and Zhang, Dehuan and Liu, Siyuan and Fu, Xianping and Li, Xuelong},
  journal={IEEE Transactions on Pattern Analysis and Machine Intelligence},
  year={2025},
  publisher={IEEE}
}

@article{wang2025fuzzy,
  title={Fuzzy Weighted Principal Component Analysis for Anomaly Detection},
  author={Wang, Sisi and Nie, Feiping and Wang, Zheng and Wang, Rong and Li, Xuelong},
  journal={ACM Transactions on Knowledge Discovery from Data},
  volume={19},
  number={3},
  pages={1--22},
  year={2025},
  publisher={ACM New York, NY}
}

@inproceedings{cao2024visible,
  title={Visible and Clear: Finding Tiny Objects in Difference Map},
  author={Cao, Bing and Yao, Haiyu and Zhu, Pengfei and Hu, Qinghua},
  booktitle=ECCV,
  year={2024}
}

@inproceedings{qiao2021detectors,
  title={Detectors: Detecting objects with recursive feature pyramid and switchable atrous convolution},
  author={Qiao, Siyuan and Chen, Liang-Chieh and Yuille, Alan},
  booktitle=CVPR,
  pages={10213--10224},
  year={2021}
}

@article{xu2022detecting,
  title={Detecting tiny objects in aerial images: A normalized Wasserstein distance and a new benchmark},
  author={Xu, Chang and Wang, Jinwang and Yang, Wen and Yu, Huai and Yu, Lei and Xia, Gui-Song},
  journal={ISPRS Journal of Photogrammetry and Remote Sensing},
  volume={190},
  pages={79--93},
  year={2022},
  publisher={Elsevier}
}

@inproceedings{yu2020scale,
  title={Scale match for tiny person detection},
  author={Yu, Xuehui and Gong, Yuqi and Jiang, Nan and Ye, Qixiang and Han, Zhenjun},
  booktitle=WACV,
  pages={1257--1265},
  year={2020}
}

@inproceedings{xia2018dota,
  title={DOTA: A large-scale dataset for object detection in aerial images},
  author={Xia, Gui-Song and Bai, Xiang and Ding, Jian and Zhu, Zhen and Belongie, Serge and Luo, Jiebo and Datcu, Mihai and Pelillo, Marcello and Zhang, Liangpei},
  booktitle=CVPR,
  pages={3974--3983},
  year={2018}
}

@article{zhu2020autoassign,
  title={Autoassign: Differentiable label assignment for dense object detection},
  author={Zhu, Benjin and Wang, Jianfeng and Jiang, Zhengkai and Zong, Fuhang and Liu, Songtao and Li, Zeming and Sun, Jian},
  journal={arXiv preprint arXiv:2007.03496},
  year={2020}
}

@inproceedings{xu2023dynamic,
  title={Dynamic coarse-to-fine learning for oriented tiny object detection},
  author={Xu, Chang and Ding, Jian and Wang, Jinwang and Yang, Wen and Yu, Huai and Yu, Lei and Xia, Gui-Song},
  booktitle=CVPR,
  pages={7318--7328},
  year={2023}
}

@inproceedings{dinh2017density,
  title={Density estimation using Real NVP},
  author={Dinh, Laurent and Sohl-Dickstein, Jascha and Bengio, Samy},
  booktitle={International Conference on Learning Representations},
  year={2017}
}

@inproceedings{li2021human,
  title={Human pose regression with residual log-likelihood estimation},
  author={Li, Jiefeng and Bian, Siyuan and Zeng, Ailing and Wang, Can and Pang, Bo and Liu, Wentao and Lu, Cewu},
  booktitle={Proceedings of the IEEE/CVF international conference on computer vision},
  pages={11025--11034},
  year={2021}
}

@inproceedings{li2017perceptual,
  title={Perceptual generative adversarial networks for small object detection},
  author={Li, Jianan and Liang, Xiaodan and Wei, Yunchao and Xu, Tingfa and Feng, Jiashi and Yan, Shuicheng},
  booktitle=CVPR,
  pages={1222--1230},
  year={2017}
}

@article{li2023dynamics,
  title={Dynamics-aware loss for learning with label noise},
  author={Li, Xiu-Chuan and Xia, Xiaobo and Zhu, Fei and Liu, Tongliang and Zhang, Xu-Yao and Liu, Cheng-Lin},
  journal={Pattern Recognition},
  volume={144},
  pages={109835},
  year={2023},
  publisher={Elsevier}
}

@article{long2024interpretable,
  title={Interpretable multidisease diagnosis and label noise detection based on a matching network and self-paced learning},
  author={Long, Jiawei and Ren, Jiangtao},
  journal={Pattern Recognition},
  volume={148},
  pages={110178},
  year={2024},
  publisher={Elsevier}
}

@article{liang2025memory,
  title={A memory-augmented multi-task collaborative framework for unsupervised traffic anomaly detection in driving videos},
  author={Liang, Rongqin and Li, Yuanman and Yi, Yingxin and Zhou, Jiantao and Li, Xia},
  journal={Pattern Recognition},
  pages={111789},
  year={2025},
  publisher={Elsevier}
}

@article{zhang2023enhanced,
  title={An enhanced noise-tolerant hashing for drone object detection},
  author={Zhang, Luming and Wang, Guifeng and Chen, Ming and Ren, Fuji and Shao, Ling},
  journal={Pattern Recognition},
  volume={143},
  pages={109762},
  year={2023},
  publisher={Elsevier}
}

@article{kanwal2024incomplete,
  title={Incomplete RGB-D salient object detection: Conceal, correlate and fuse},
  author={Kanwal, Samra and Taj, Imtiaz Ahmad},
  journal={Pattern Recognition},
  pages={110700},
  year={2024},
  publisher={Elsevier}
}

@inproceedings{OAMIL,
  title={Robust Object Detection with Inaccurate Bounding Boxes},
  author={Liu, Chengxin and Wang, Kewei and Lu, Hao and Cao, Zhiguo and Zhang, Ziming},
  booktitle={European Conference on Computer Vision},
  pages={53--69},
  year={2022},
  organization={Springer}
}

@inproceedings{SDlocnet,
  title={Learning to localize objects with noisy labeled instances},
  author={Zhang, Xiaopeng and Yang, Yang and Feng, Jiashi},
  booktitle={Proceedings of the AAAI Conference on Artificial Intelligence},
  volume={33},
  number={01},
  pages={9219--9226},
  year={2019}
}

@inproceedings{chadwick2019training,
  title={Training object detectors with noisy data},
  author={Chadwick, Simon and Newman, Paul},
  booktitle={2019 IEEE Intelligent Vehicles Symposium (IV)},
  pages={1319--1325},
  year={2019},
  organization={IEEE}
}

@article{li2020towards,
  title={Towards noise-resistant object detection with noisy annotations},
  author={Li, Junnan and Xiong, Caiming and Socher, Richard and Hoi, Steven},
  journal={arXiv preprint arXiv:2003.01285},
  year={2020}
}

@article{chen2025hieratok,
  title={HieraTok: Multi-Scale Visual Tokenizer Improves Image Reconstruction and Generation},
  author={Chen, Cong and Huang, Ziyuan and Zou, Cheng and Zhu, Muzhi and Ji, Kaixiang and Liu, Jiajia and Chen, Jingdong and Chen, Hao and Shen, Chunhua},
  journal={arXiv preprint arXiv:2509.23736},
  year={2025}
}

@article{shi20252d,
  title={2D Gaussians Meet Visual Tokenizer},
  author={Shi, Yiang and Guo, Xiaoyang and Yin, Wei and Jia, Mingkai and Zhang, Qian and Hu, Xiaolin and Liu, Wenyu and Wang, Xinggang},
  journal={arXiv preprint arXiv:2508.13515},
  year={2025}
}
\end{document}